\documentclass[10pt,twocolumn,letterpaper]{article}

\usepackage[pagenumbers]{cvpr} 
\usepackage[table]{xcolor}

\usepackage{multirow}
\usepackage{bm}
\usepackage{amsmath}

\usepackage{adjustbox}

\definecolor{cvprblue}{rgb}{0.21,0.49,0.74}
\usepackage[pagebackref,breaklinks,colorlinks,allcolors=cvprblue]{hyperref}
\usepackage{amsmath}  
\usepackage{amsthm}

\def\paperID{17065} 
\def\confName{CVPR}
\def\confYear{2026}

\newcommand{\datasetname}{GarMoCap} 

\title{Garment Inertial Denoiser: Endowing Accurate Motion Capture via Loose IMU Denoiser}

\author{
Jiawei Fang$^{1}$\thanks{Equal contribution.} \quad
Ruonan Zheng$^{1}$\footnotemark[1] \quad
Xiaoxia Gao$^{1}$\footnotemark[1] \quad
Shifan Jiang$^{1}$ \quad
Anjun Chen$^{1}$\thanks{Corresponding author} \quad
Qi Ye$^{2}$ \quad
Shihui Guo$^{1}$ \\
$^{1}$Xiamen University \quad
$^{2}$Zhejiang University
}

\begin{document}
\maketitle
\begin{abstract}
Wearable inertial motion capture (MoCap) provides a portable, occlusion-free, and privacy-preserving alternative to camera-based systems, but its accuracy depends on tightly attached sensors—an intrusive and uncomfortable requirement for daily use. Embedding IMUs into loose-fitting garments is a desirable alternative, yet sensor–body displacement introduces severe, structured, and location-dependent corruption that breaks standard inertial pipelines. We propose \textbf{GID} (Garment Inertial Denoiser), a lightweight, plug-and-play Transformer that \emph{factorizes} loose-wear MoCap into three stages: (i) location-specific denoising, (ii) adaptive cross-wear fusion, and (iii) general pose prediction. GID uses a \emph{location-aware expert} architecture, where a shared spatio-temporal backbone models global motion while per-IMU expert heads specialize in local garment dynamics, and a lightweight fusion module ensures cross-part consistency. This inductive bias enables stable training and effective learning from limited paired loose–tight IMU data. We also introduce \textbf{GarMoCap}, a combined public and newly collected dataset covering diverse users, motions, and garments. Experiments show that GID enables accurate, real-time denoising from single-user training and generalizes across unseen users, motions, and garment types—consistently improving state-of-the-art inertial MoCap methods when used as a drop-in module.
\end{abstract}

\section{Introduction}
Human pose estimation is a fundamental task with broad applications in healthcare, sports, and immersive interaction~\cite{zheng2023deep}. As an alternative to traditional vision-based methods, wearable motion capture using Inertial Measurement Units (IMUs) offers compelling advantages in portability, privacy-friendliness, and robustness against occlusion or extreme lighting~\cite{filippeschi2017survey}. Recent work has demonstrated impressive full-body tracking using only a sparse set (3-6) of IMUs~\cite{huang2018deep,yi2021transpose,yi2022physical,yi2024physical,wu2024accurate,zuo2025transformer}. However, these systems typically rely on IMUs being tightly attached to the body to ensure stable inertial readings, which can be intrusive and uncomfortable for daily or long-term use, such as human data collection, chronic disease monitoring, and daily body-centric interaction. An ideal solution, which would largely reduce the burden on users, is to integrate IMUs directly into everyday, loose-fitting clothing.
\begin{figure}[t]
    \centering
    \includegraphics[width=\linewidth]{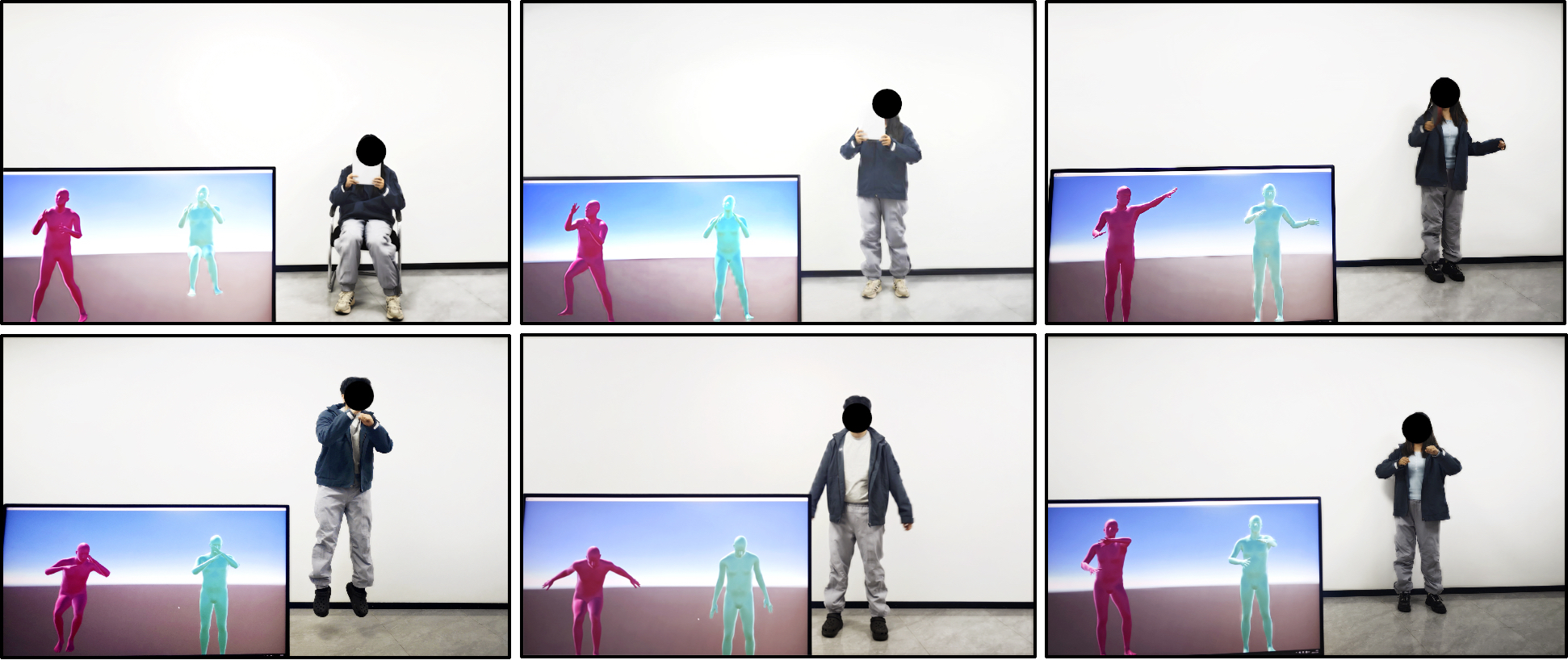}
    \caption{
        \textbf{Garment Inertial Denoiser (GIP) can denoise real-time loose wearing IMU data, enabling robust full-body motion capture. }
    }
    \label{fig:teaser}
\end{figure}

However, this loose-garment setup introduces a critical challenge: the secondary motion between the garment and the body (e.g., slippage, swinging) severely corrupts the IMU data. A common strategy to handle such data shifts is domain randomization~\cite{tobin2017domain}, which trains models on data augmented with stochastic perturbations to promote robustness. Nevertheless, we argue that this approach is fundamentally mis-specified for this problem. The noise induced by garment dynamics is inherently structured rather than i.i.d., characterized by high-dimensional, non-stationary, and body-part-dependent variations. Expanding the augmentation space to capture this complex distribution leads to an explosion in hypothesis complexity, unstable optimization, and inflates data requirements, while ultimately failing to cover real-world structured shifts.

\begin{figure*}[t]
    \centering
    \includegraphics[width=\linewidth]{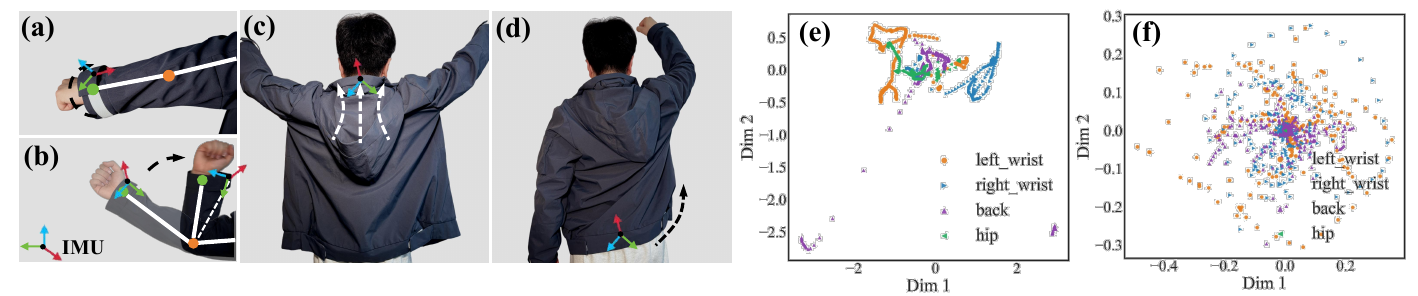}
\caption{
\textbf{Motivation.}
(a–d) Loose-wear garments introduce location-dependent disturbances to IMU signals, including fabric sliding, impact, and garment deformation. 
(e) PCA of \emph{rotation} signals shows that tight-wear IMUs form clean, body-part–specific manifolds, 
whereas (f) PCA of \emph{acceleration} signals reveals large spread and collapsed structure under loose-wear conditions. 
These observations highlight the need for our GID module to recover tight-wear signal characteristics from loose-wear measurements.
}

    \label{fig:motivation}
\end{figure*}

Therefore, instead of learning a direct mapping from corrupted inertial signals to body pose, we propose a \emph{factorized learning framework} to decouple garment-induced corruption from pose estimation. 
Specifically, we decompose the overall mapping 
$f_{\mathbf{IMU}_{\text{loose}} \rightarrow \boldsymbol{\theta}}$ 
into two sub-functions:
\begin{equation}
    f_{\mathbf{IMU}_{loose} \rightarrow \boldsymbol{\theta}}
    =
    f_{\mathbf{IMU}_{loose} \rightarrow \mathbf{IMU}_{tight}}
    \circ
    f_{\mathbf{IMU}_{tight} \rightarrow \boldsymbol{\theta}}
    .
\end{equation}
Here,
$f_{\mathbf{IMU}{tight} \rightarrow \boldsymbol{\theta}}$
denotes the well-established, data-rich pose estimator trained on tight-wear IMU signals, while
$f_{{\mathbf{IMU}{loose}} \rightarrow {\mathbf{IMU}_{tight}}}$
is our proposed \emph{Garment Inertial Denoiser (GID)}, which maps loose-wear IMU measurements to their tight-wear counterparts (Figure~\ref{fig:teaser}).
This factorization explicitly isolates garment-induced disturbances from the downstream pose prediction task, significantly reducing the effective learning complexity when only limited paired loose–tight data are available. As a result, it improves both data efficiency and optimization stability. In practice, however, directly training
$f_{\mathbf{IMU}{loose} \rightarrow \mathbf{IMU}{tight}}$
as a single end-to-end network is ineffective. Loose-wear IMU corruption is highly heterogeneous and location-dependent: sensors on different body segments (e.g., wrist vs. torso) exhibit distinct garment–body dynamics. When modeled within a single monolithic denoiser, these heterogeneous patterns introduce conflicting optimization and lead to unstable learning (Figure~\ref{fig:motivation}).

To address this challenge, GID adopts a position-aware expert architecture. A shared spatio-temporal Transformer backbone captures global motion regularities, while lightweight per-IMU expert heads specialize in the local physics associated with their respective sensor locations. A small fusion module further enforces cross-part consistency and resolves inter-expert discrepancies. This hierarchical design embeds the right inductive biases and enables reliable learning from minimal paired data, while generalizing effectively across users and garment configurations.

To evaluate our approach under realistic design shifts, we introduce GarMoCap, a comprehensive dataset for garment-based motion capture, comprising 4 different garment designs with 43 users across 10 expressive motions. Experimental results show that GID trained using data from a single user wearing a single garment can generalize to unseen users and garment designs effectively. Serving as a plug-and-play module, GID consistently reduces angular and positional errors and enhances the robustness of existing pose estimation models without requiring any architectural modification.
The contributions of our work can be summarized as follows: 

\begin{itemize}
\item We reformulate loose-wear inertial MoCap as a structured denoising task and propose a \textit{factorized} framework that isolates garment-induced artifacts from pose prediction, allowing direct reuse of existing tight-wear IMU models.
\item We propose the Garment Inertial Denoiser (GID), a lightweight, location-specific denoising module that models body-part-specific garment-induced motion artifacts. By assigning separate denoisers to each IMU channel, GID mitigates conflicts in learning dynamics across heterogeneous sensor locations and improves training stability.
\item We present \textbf{GarMoCap}, a new dataset featuring diverse users, motions, and garment designs, specifically constructed to study real-world garment–body interaction and loose-wear sensor corruption.
\item We show that GID is highly data-efficient—training on a single garment—on a single user yet generalizes to unseen users and garment types, and functions as a drop-in module that significantly improves the accuracy and robustness of existing inertial MoCap pipelines.
\end{itemize}
\section{Related Work}

\subsection{Motion Capture}
\subsubsection{Vision-based MoCap}
Optical approaches fall into marker-based~\cite{chatzitofis2021democap,ghorbani2021soma,Optitrack2024} and marker-less~\cite{Rahul2018} categories. Commercial systems such as Vicon~\cite{Vicon2024} deliver high fidelity but require costly infrastructure and tightly controlled studios, which limits everyday accessibility. Among marker-less methods, multi-camera systems~\cite{choudhury2023tempo,liao2024multiple,wu2021graph,zhang2021direct} use calibrated, synchronized camera arrays with triangulation and volumetric inference to substantially reduce occlusions and improve accuracy, but they still demand careful calibration, dense hardware, and significant compute, making deployment space- and cost-prohibitive in many settings~\cite{mehta2020xnect}. In contrast, single-camera RGB/RGB-D pipelines~\cite{sun2019deep,zhen2020smap,kocabas2021pare,xu2022vitpose,yang2021transpose,zhao2023single,shin2024wham,feng2023diffpose,xia2025reconstructing} have gained traction for their simplicity and scalability, yet remain sensitive to occlusions, lighting, clothing appearance, and typically assume static cameras—reducing robustness in outdoor or highly dynamic scenes. These constraints still hinder broad, in-the-wild deployment of purely vision-based MoCap.
\subsubsection{Inertial-based MoCap}
Inertial Measurement Units (IMUs) provide a portable and lighting-invariant alternative to vision-based motion capture by directly measuring segment orientations and linear accelerations. Commercial systems such as Noitom~\cite{Noitom2024} and Xsens~\cite{Movella2024} achieve high accuracy by tightly strapping dense arrays of IMUs to many body segments, but these setups are intrusive, restrict natural movement, and are expensive to deploy. These limitations have motivated a broad shift toward sparse IMU configurations (typically 3–6 sensors) that aim to balance practicality with fidelity.

Early work demonstrated the feasibility of sparse IMU pose reconstruction: SIP~\cite{von2017sparse} used offline optimization, while DIP~\cite{huang2018deep} achieved near real-time performance with a bidirectional RNN but focused primarily on local pose. TransPose~\cite{yi2021transpose} made a key step forward by introducing a multi-stage network and modeling foot–ground interactions to recover global translation and reduce drift.
Subsequent research has advanced along two directions: enforcing stronger physical consistency and designing better network architectures. PIP~\cite{yi2022physical} introduced physics-based regularization, and PNP~\cite{yi2024physical} further improved robustness by explicitly modeling fictitious forces in non-inertial frames. On the architectural side, TIP~\cite{jiang2022transformer} brought Transformers into sparse IMU pose estimation, and ASIP~\cite{wu2024accurate} combined a Transformer backbone with adversarial training to enhance physical plausibility.
More recently, system usability has also become an important focus. TIC~\cite{zuo2025transformer} proposed a dynamic on-body calibration strategy that removes the reliance on static T-pose calibration, significantly improving long-term stability and user convenience.
Concurrently, multimodal fusion has emerged as another direction for enhancing robustness. This includes fusing UWB~\cite{armani2024ultra,devrio2023smartposer}, electromagnetic (EM) fields~\cite{kaufmann2021pose} to improve skeletal tracking and global displacement estimation. Other works, particularly in AR/VR scenarios, focus on extrapolating full-body pose from sparse inputs, such as from the upper body or just head and hands~\cite{du2023avatars,jiang2022avatarposer,yang2024divatrack,ponton2023sparseposer}.

Although innovations in sparse IMU systems have alleviated some discomfort associated with dense configurations, their fundamental reliance on tight-wearing straps persists. This requirement for secure fixation severely impacts long-term user comfort, posing a challenge to sustained wearability and practicality in daily, extended-use scenarios.

\subsection{Clothes-based MoCap}
Integrating motion capture (MoCap) into everyday clothing offers a promising path toward long-term, portable, and privacy-preserving tracking~\cite{fadhil2023human}. Existing garment-based approaches can be broadly divided by how the clothing fits. Tight-wear systems maintain fixed sensor–body alignment, enabling stable and accurate measurements. Representative examples include dense IMU arrays~\cite{fadhil2023human}, flexible joint-tracking sensors~\cite{chen2023dispad,jiawei2024suda}, posture-recognition textiles~\cite{liang2023smart}, and hybrid systems combining sparse depth cameras with optical markers~\cite{chatzitofis2021democap}. SuDA~\cite{jiawei2024suda} further introduced a support-based domain adaptation strategy to address signal instability caused by garment deformation during wear.

While form-fitting garments provide high data accuracy, loose-wear options better suit everyday comfort needs. Research in this area has concentrated primarily on IMUs and capacitive sensors. Regarding capacitive sensors, Zhou et al. ~\cite{zhou2023mocapose} embedded multi-channel sensors into a jacket, employing a deep regressor for upper body coordinate estimation. SeamPose~\cite{yu2024seampose} involved placing sensors covertly within the seams of clothing, which led to SeamFit~\cite{yu2025seamfit}, a system of washable, multi-sized T-shirts that can infer the movement of joints not directly covered. In the IMU domain, Zhou et al.~\cite{lorenz2022towards} investigated the use of multiple IMUs to translate data from loose-wear to a tight-wear equivalent. LIP~\cite{zuo2024loose} embedded 4 sparse IMUs in a loose jacket and introduces a Secondary Motion Autoencoder (SeMo-AE) that models garment–skin secondary motion as additive Gaussian noise in a learned latent space and, together with a temporal-coherence scheme, synthesizes diverse loose-wear IMU signals from tight-wear data to augment training. Other work has limited the use of clothing-mounted IMUs to activity recognition~\cite{jayasinghe2019comparing,michael2017activity,shen2023probabilistic}. Recent investigations are exploring clothing as a platform for multimodal sensor fusion to gain synergistic benefits. For instance, FIP~\cite{zheng2025fip} fused 4 IMUs and 2 elbow flex sensors in a loose-fitting daily jacket and introduces a Displacement Latent Diffusion Model to synthesize and resist IMU real-time displacement, enabling robust, real-time clothes-based MoCap. Most recently, Ilic et al.~\cite{ilic2025human} introduced GaIP, a garment-aware diffusion framework that estimates full-body pose from sparse, loosely attached IMUs by synthesizing loose-wear signals and conditioning on clothing parameters to handle sensor–garment dynamics.

However, SOTA sparse IMU systems like LIP and FIP algorithmically rely heavily on data-driven noise modeling which is time cosuming and need massive training data in an attempt to cover all possible noise conditions which is nearly impossible. To address these limitations, we propose the first full-body loose-wear MoCap system based on sparse IMUs (n=6). Instead of relying on large-scale data synthesis, our method requires only a small amount of real data to train a lightweight, plug-and-play denoiser, enhancing both data efficiency and training stability.

\begin{figure*}[t]
    \centering
    \includegraphics[width=\linewidth]{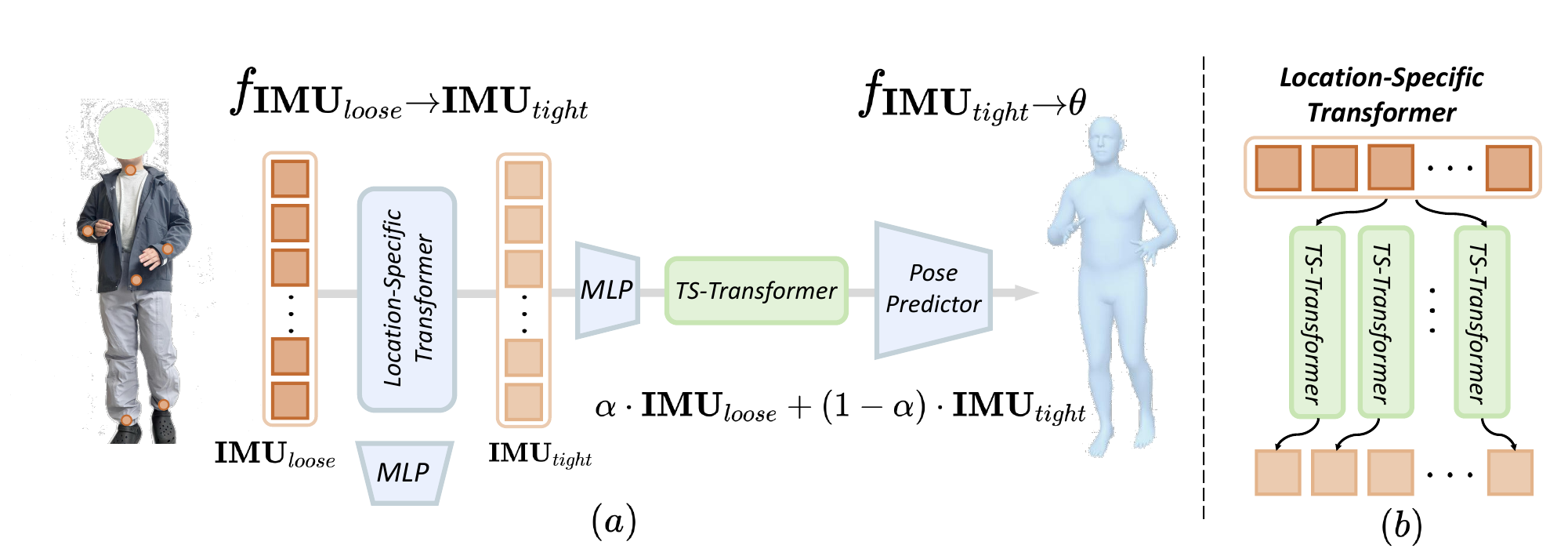}
    \caption{
        \textbf{Garment Inertial Denoiser pipeline.} (a) We factorize $f$ into two sub-functions corresponding to signal denoising and pose reconstruction. (b) Each sensor reading $\mathbf{IMU}_{loose}$ will be assigned to a Temporal-Spatial(TS) transformer-based location-specific denoiser. 
    }
    \label{fig:pipeline}
\end{figure*}


\section{Method}
\subsection{Problem Definition} 
Let $\bm \theta \in \mathbb R^{J \times3}$ represent the 3D rotation angles for $J$ body joints. The in-garment IMU sensor network consists of $M$ IMUs located at $\mathbf s_m, m =1,2,3,...M$. Each IMU  $m$ captures acceleration $\mathbf{a}_m \in \mathbb R^{3}$ and rotation  $\mathcal{R}_m \in \mathbb R^{3}$ at its corresponding location. In this work, we aim to develop a garment-based motion capture system using sparse inertial sensors to reconstruct the mapping function $f$ between the loose IMU readings $\mathbf{IMU}_{loose}$ and the underlying joint rotations $\bm{\theta}$:
\begin{equation}
    \bm {\theta} = f(\mathbf{IMU}_{loose})
    \label{eq:function}
\end{equation}

However, the direct estimation of $f$ is challenging. Loose-wear IMUs introduce secondary motions arising from fabric–body displacement, which manifest as complex disturbances overlaid on the ideal tight-wear signals. These disturbances substantially expand the parameter space of the mapping function $f$ (Eq.~\ref{eq:function}), thereby complicating direct optimization. Inspired by domain randomization~\cite{tobin2017domain}, a straightforward method is to synthesize a variety of loose-wear IMU data by adding noise into tight-wear signals. The model $f$ is then trained on this augmented data, with the expectation that it will generalize to real-world $\mathbf{IMU}_{\text{loose}}$ conditions:
\begin{equation}
    f_{\mathbf{IMU}_{loose} \to \bm \theta} =  f_{\mathbf{IMU}_{tight} \to \mathbf{IMU}_{loose}} \circ f_{\mathbf{IMU}_{loose} \to \bm \theta} 
\end{equation}

This strategy, however, is fundamentally mis-specified for the problem of garment motion. The disturbance from loose clothing is not i.i.d. noise but a structured physical perturbations posed by physical properties such as inertia, momentum, and material properties. The noise injection approach ignores this underlying structure, attempting to achieve robustness by brute-forcing a massive parameter space. This leads to an inherent dilemma: narrow noise distributions fail to cover real-world dynamics, while broad distributions destabilize training and learn a mapping that is both computationally inefficient and physically implausible. 

\subsection{Garment Inertial Denoiser}

\paragraph{Factorized Parameter Space.}
To overcome these limitations, we adopt a different perspective: instead of randomizing noise, we explicitly learn a denoising function $f_{\mathbf{IMU}_{loose} \to \mathbf{IMU}_{tight}}$ that maps loose-wear IMU readings to their tight-wear counterparts. The key insight underlying our formulation is that the parameter space of $f$ is \textit{factorizable}, allowing it to be decomposed into two interpretable sub-functions corresponding to signal denoising and pose reconstruction:
\begin{equation}
    f_{\mathbf{IMU}_{loose} \to \bm \theta} =  f_{\mathbf{IMU}_{loose} \to \mathbf{IMU}_{tight}} \circ f_{\mathbf{IMU}_{tight} \to \bm \theta} 
    \label{eq:denoise}
\end{equation}
This decomposition strategy is advantageous. It constrains the hypothesis space and enables more efficient learning from limited paired data.
Since the mapping $f_{\mathbf{IMU}_{tight}\rightarrow \boldsymbol{\theta}} $ encapsulates complex body dynamics and non-linear kinematic dependencies, which occupy a substantially larger portion of the parameter space. In contrast, In contrast, the denoising mapping ($f_{\mathbf{IMU}_{loose} \to \mathbf{IMU}_{tight}}$) is a comparatively lower-complexity problem as $\mathbf{IMU}_{loose}$ and $\mathbf{IMU}_{tight}$ share similar spatial and temporal structures. The task is not to learn pose from scratch, but to learn the inverse of this structured physical disturbance. This factorization allows us to decouple the problem accordingly: we use scarce paired data to learn the disturbance model, while leveraging mature, data-rich estimators for the pose reconstruction task.

\paragraph{Location-Specific Denoising.}
Nonetheless, directly learning the denosing function $f_{\mathbf{IMU}_{loose}\to\mathbf{IMU}_{tight}}$ in Eq.~\ref{eq:denoise} is highly non-trivial. The main challenge lies in the \emph{location-dependent} nature of loose-wear corruption (Figure~\ref{fig:motivation}). IMUs mounted at different body sites (\textit{e.g.}, wrist, back, waist) are subject to distinct joint kinematics and sensor–body–garment coupling dynamics, shaped by factors such as attachment geometry, local slack, and garment–skin contact patterns. Consequently, the resulting signal statistics vary systematically across locations. Using a single shared model to represent such heterogeneous distributions introduces conflicting optimization gradients, making it difficult to learn an effective and generalizable mapping.

To address this, GID employs a \textit{Location-Specific Denoising} paradigm, in which each sensor is assigned a location-specific denoiser (expert module) that models its unique physical behavior. These experts operate under different spatio-temporal Transformer backbones that captures global motion regularities, while a lightweight fusion module enforces cross-part consistency. This design embeds appropriate inductive biases, enhances robustness to heterogeneous motion artifacts, and maintains parameter efficiency, making the composite mapping in Eq.~\ref{eq:denoise}
both tractable and generalizable.

\paragraph{Adaptive Cross-wear Fusion.}
While each location-specific denoiser produces a refined estimation of $\hat{\mathbf{IMU}}_{tight}$, their outputs may still exhibit residual inconsistencies due to independent processing. To reduce these discrepancies and leverage complementary cues from both tight- and loose-wear domains, we introduce an \emph{early fusion} strategy. Specifically, we construct a learnable convex combination as
$\alpha \cdot \mathbf{\hat 
{IMU}}_{tight} + (1-\alpha)\cdot\mathbf{IMU}_{loose}$,
where $\alpha\in[0,1]$ is a trainable weight realized via a sigmoid parameterization. This formulation allows the model to adaptively balance the stability of tight-wear signals and the realism of loose-wear dynamics, capturing both high-frequency and low-frequency motion patterns. 
The fused representation is then propagated through our final temporal-spatial transformer for consistency refinement, yields smoother pose reconstruction without additional inference overhead.

\paragraph{Pose Predictor.}
The resulting filtered tight-equivalent signals are subsequently fed into the pose predictor 
$f_{\mathbf{IMU}_{\text{tight}}\rightarrow \boldsymbol{\theta}}$, 
which maps inertial sequences to the corresponding body pose parameters $\boldsymbol{\theta}$. 
Given the abundance of paired tight-wear IMU and motion data, this stage can be trained under fully supervised settings, enabling precise recovery of body kinematics. 
Thanks to its modular and lightweight formulation, our pipeline remains agnostic to the specific instantiation of 
$f_{\mathbf{IMU}_{\text{tight}}\rightarrow \boldsymbol{\theta}}$ 
and can seamlessly integrate with SOTA IMU-to-pose architectures (\textit{e.g.}, PIP, PNP) without structural modification. 
This decoupled design isolates the location-aware denoisers from the downstream pose predictor, facilitating flexible adaptation, stable convergence, and efficient learning from limited paired data.



\section{\datasetname ~Dataset}
To comprehensively evaluate our method, we construct \datasetname, a composite dataset collection that integrates existing public upper body datasets (15 users) with our newly collected data (28 users) with full body MoCap garment (see Figure~\ref{fig:hardware}). This collection covers diverse garment types ($n=4$), body regions ($n=43$), with both upper-body and full-body recordings.

\subsection{Hardware}

\begin{figure}[t]
    \centering
    \includegraphics[width=\linewidth]{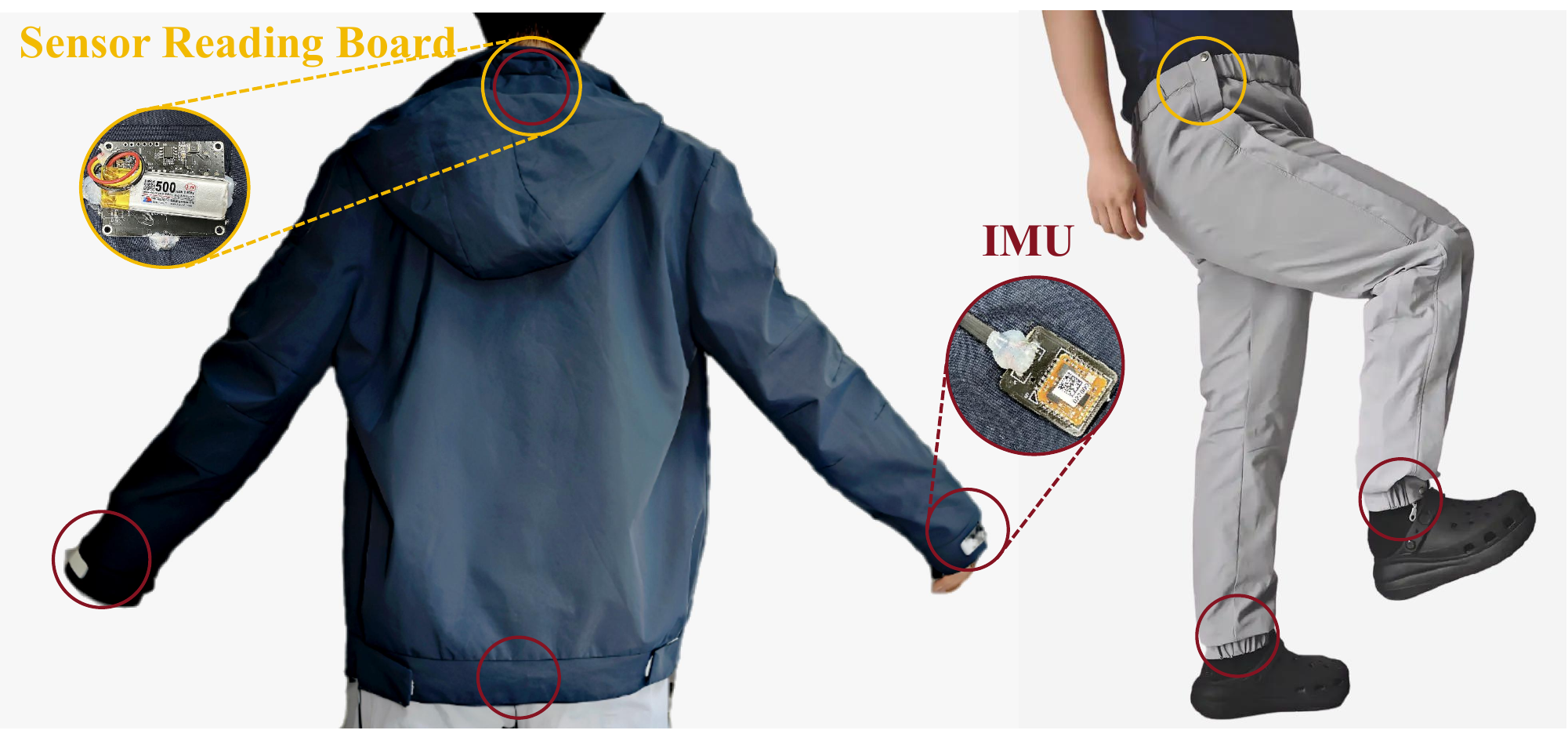}
    \caption{
        \textbf{Prototype of our full-body garment-based motion capture system.}
        The loose-wear jacket and pants are embedded with six IMU sensors and two central circuit boards for data acquisition in total.
    }
    \label{fig:hardware}
\end{figure}

Figure~\ref{fig:hardware} illustrates our prototype outfit for full-body MoCap built on everyday sportswear patterns. The size of the garment is XL, which is designed with target fit metrics of 50 cm (shoulder width), 108 cm (chest), 18 cm (waist), and 25 cm (sleeve circumference), intended to accommodate wearers ranging from 160–185 cm in height and 50–85 kg in weight.
The system consists of a jacket and trousers, each integrating four Xsens MTI-3 IMUs~\cite{xsens2025}. The sensor placements on the upper garment are left forearm, right forearm, back (co-located with the data-transfer mainboard), and waistline. Placements on the lower garment include left waist and right waist (integrating the lower mainboard). As shown in technical outerwear, each sensor is connected to its respective mainboard via flexible cables. These cables are heat-bonded into the fabric to ensure a flush, durable, and unnoticeable routing. Both mainboards sample their connected IMUs at 40 Hz and relay the measurements wirelessly to a host computer via Bluetooth. The system is powered by a 500 mAh lithium battery. For more information, please refer to our appendix.

\subsection{Dataset Collection}
Using our garment MoCap system, we collect two new datasets, $\mathcal D_{GID}^{upper}$ and $\mathcal D_{GID}^{full}$. We then integrate them with two existing public garment MoCap datasets, $\mathcal D_{FIP}^{upper}$ and $\mathcal D_{LIP}^{upper}$, to form our composite data collection \datasetname.

\begin{itemize}
    \item $\mathcal D_{GID}^{upper}$: We collect the \textbf{GID-Upper} dataset, consisting of twenty participants (10 males, 10 females) wearing the loose-fit jacket instrumented with four IMUs. Participants perform five sequential upper-body actions (\textit{e.g.,} running, boxing, warm-up) each for one minute, totaling approximately 100 minutes of data. Ground truth is recorded via the Perception Neuron 3 inertial system~\cite{perceptionneuron2025}. This dataset is specifically designed to evaluate cross-user and cross-motion generalization.

    \item $\mathcal D_{GID}^{full}$: We further collect the \textbf{GID-Full} dataset, containing recordings from eight participants performing ten full-body actions (approximate 80 minutes total). This data is captured using our full-body garment set (jacket and pants) with integrated 6 IMUs. Notably, ground truth is obtained using an 8-camera synchronized RGB array, as traditional optical markers are infeasible when obscured by loose clothing. The cameras cover a 4m$\times$2.5m capture volume, allowing unrestricted movement. We utilize the EasyMocap~\cite{easymocap} framework for 3D reconstruction of SMPL~\cite{loper2023smpl} parameters (shape $\beta$, joint rotations, translation), achieving a reconstruction error below 20 mm. This multi-camera setup effectively mitigates occlusion, ensuring reliable ground truth generation even for dynamic, loose-wear scenarios. 

    \item $\mathcal D_{FIP}^{upper}$: We leverage the \textbf{FIP-Upper} dataset~\cite{zheng2025fip}, a collection of real-world recordings for evaluating IMU garments. It features ten participants with diverse body shapes performing ten predefined upper-body actions (\textit{e.g.,} walking, running, boxing). The dataset contains 371,122 frames captured at 60~fps (approx. 103 minutes), and ground-truth poses are also provided by a Perception Neuron 3 system using eleven tightly-worn IMUs.

    \item $\mathcal D_{LIP}^{upper}$: We also incorporate the \textbf{LIP-upper} dataset~\cite{zuo2024loose}, designed for evaluating loose-wear IMU garments. It includes five participants wearing a jacket with four IMUs under two conditions (zipped and unzipped) to simulate varying looseness. The dataset comprises 212,496 frames at 30~fps (approx. 118 minutes) covering both predefined actions and free-form movements. Ground truth is also captured by a Perception Neuron 3 system with eleven tight IMUs.

\end{itemize}

Together, these datasets form a unified evaluation corpus covering diverse garment designs, wearing conditions (zipped, unzipped), and body coverage (upper and full body). This collection provides a comprehensive foundation for studying \emph{loose–tight domain translation}, \emph{sensor disturbance denoising}, and \emph{cross-garment generalization} in wearable motion capture. Please refer to our supplementary materials for details on our GarMoCap.

\section{Experiment}

\subsection{Experiment Setup}

\paragraph{Metrics}
We measure the denoising performance with our method using the MAE (Mean Absolute Error) loss. Following TransPose~\cite{yi2021transpose}, we measure the accuracy of pose estimation using the following four metrics: 

\begin{itemize}
    \item Angular Error, which measures the mean global rotation error of all body joints in degrees.
    \item Positional Error, which measures the mean Euclidean distance error of all estimated joints in centimeters.
    
    \item Mesh Error, which measures the vertex error of the posed SMPL meshes.
    \item Jitter Error, which measures the mean jerk (time derivative of acceleration) of all body joints in the global space, which reflects the smoothness of the motion~\cite{flash1985coordination}. 
\end{itemize}

\paragraph{Training Details} All experiments were performed on a workstation equipped with an Intel U9-285K CPU and an NVIDIA RTX 5080 GPU. The implementation is based on PyTorch 2.8.0 with CUDA 12.9. Additional training configurations and network specifications are included in the supplementary material.

\subsection{Performance on the Same Garment Design}

\paragraph{Full-body MoCap.}
We first evaluate GID on full-body motion capture, using one subject from $\mathcal{D}_{GID}^{full}$ for training and the remaining subjects for testing. Full-body MoCap is substantially more challenging than upper-body tracking due to its higher-dimensional parameter space and the strong inter-limb dependencies that propagate garment-induced noise across joints. Despite these difficulties, GID delivers consistently large improvements across all protocols—PIP and ASIP in particular—demonstrating that our denoiser effectively removes garment motion and restores physically meaningful inertial signals even under complex whole-body dynamics (Table~\ref{tab:full_body}).

\begin{table}[t]
    \centering
    \caption{\textbf{Comparison of SOTA methods with and without GID on full-body performance.} Lower values indicate better performance.}
    \vspace{1mm}
    \resizebox{\linewidth}{!}{
    \begin{tabular}{lcccc}
        \toprule
        \textbf{Method} & \textbf{Ang. (°)} & \textbf{Pos. ($cm$)} & \textbf{Mesh ($cm$)} & \textbf{Jitter ($m/s^3$) } \\
        \midrule
        PIP~\cite{yi2022physical}& \textbf{16.58}/24.27 & \textbf{8.95}/10.85 & \textbf{10.79}/13.01 & \textbf{0.37}/0.40 \\
        PNP~\cite{yi2024physical}  & \textbf{23.21}/23.21 & \textbf{10.95}/11.94 & \textbf{13.28}/14.05 & 1.03/\textbf{0.86} \\
        ASIP~\cite{wu2024accurate}  & \textbf{15.68}/26.09 & \textbf{8.49}/11.49 & \textbf{10.45}/14.02 & \textbf{0.03}/0.07 \\
        \bottomrule
    \end{tabular}
    }
    \label{tab:full_body}
\end{table}
One interesting observation is that applying GID-denoised IMU signals to PNP does not further improve its performance. We attribute this to a representational mismatch: our denoiser operates in a root-relative IMU space, whereas PNP relies on world-frame, physically structured inertial cues. Injecting root-relative denoised signals into a world-frame predictor may distort the temporal dynamics that PNP depends on. Training GID directly in world coordinates, however, would re-introduce camera alignment and subject-specific global motion into the learning target, making the task significantly harder and empirically leading to unstable optimization and poorer generalization.

\begin{table}[t]
    \centering
    \caption{\textbf{Comparison of SOTA MoCap methods with / without GID on upper-body dataset evaluation.}}
    \resizebox{\linewidth}{!}{
    \begin{tabular}{lcccc}
        \toprule
        \textbf{Method} &
        \textbf{Ang. (°)} & \textbf{Pos. ($cm$)} & \textbf{Mesh ($cm$)} & \textbf{Jitter ($m/s^3$) } \\
        \midrule
        PIP  & {\;\;\textbf{38.24}/38.42\;\;} & 16.08/\textbf{14.63} & 18.84/\textbf{17.97} & \textbf{0.34}/0.57 \\
        PNP  & {\;\;\textbf{20.39}/22.38\;\;} & \textbf{8.80}/9.22 & \textbf{11.36}/13.17 & \textbf{0.35}/0.63 \\
        ASIP & {\;\;\textbf{20.08}/24.18\;\;} & \textbf{9.07}/10.13 & \textbf{11.07}/13.69 & \textbf{0.13}/0.30 \\
        \bottomrule
    \end{tabular}
    }
    \label{tab:upper}
\end{table}

\paragraph{Upper-body MoCap.}  

To assess the intra-garment generalization ability of GID, we evaluate on the upper-body subset extracted from the full-body recordings. Since the garment top can be worn independently, this setting directly tests whether GID-trained signals remain useful outside the full-body context. We train on one randomly selected user from $\mathcal{D}_{GID}^{full}$ and test on the remaining eight. As shown in Table~\ref{tab:upper}, despite using paired loose–tight IMU data from only a single subject, our denoiser reliably suppresses loose-IMU artifacts and yields substantial improvements in pose estimation accuracy. These gains mirror our full-body results, further demonstrating that GID generalizes effectively across users and wearable configurations.


\subsection{Generalization to Different Garment Designs}

To further assess the cross-garment generalization ability of GID, we used one user's data from $\mathcal{D}_{GID}^{full}$ for training and evaluated the model on $\mathcal{D}_{LIP}^{upper}$, $\mathcal{D}_{FIP}^{upper}$, and $\mathcal{D}_{GID}^{upper}$.  
Notably, these datasets together cover three distinct garment designs and a total of 35 users.  
As shown in Table~\ref{tab:all_protocols_gid_smallIsBold}, despite being trained on a single user, GID demonstrates robust generalization, maintaining high denoising accuracy and, consequently, reliable motion capture performance across garments of varied materials, patterns, and looseness.  
This finding aligns well with our core hypothesis that the GID representation captures garment-invariant motion features.

\subsection{Comparison with SOTA Garment-based MoCap Methods}

We compare GID with two SOTA garment-based MoCap methods, FIP and LIP, using the official checkpoints from their public repositories. Both baselines perform strongly on $\mathcal{D}_{LIP}^{upper}$, $\mathcal{D}_{FIP}^{upper}$, and $\mathcal{D}_{GID}^{upper}$ (Table~\ref{tab:method_dataset_comparison}). Notably, despite being trained on data from a single garment worn by one participant (38,286 frames; 11 min), GID matches or exceeds their accuracy, whereas FIP/LIP rely on over 7 million simulated frames (1,967 min), highlighting a clear advantage in data efficiency and training cost. In addition, GID supports full-body MoCap, while FIP/LIP target upper body only; accordingly, we report quantitative comparisons on upper-body datasets.

\begin{table*}[t]
    \centering
    \caption{\textbf{Comparison of SOTA MoCap methods with and without GID across $\mathcal D_{LIP}^{upper}$, $\mathcal D_{FIP}^{upper}$, and $\mathcal D_{GID}^{upper}$ datasets.} Despite being trained on data from a single garment worn by a single user, GID generalizes effectively to unseen domain with garment designs, users, and motions. }
    \vspace{1mm}
    \resizebox{\linewidth}{!}{
    \begin{tabular}{lcccccccccccc}
        \toprule
        \multirow{2}{*}{\textbf{Method}} &
        \multicolumn{4}{c}{$\mathcal D_{LIP}^{upper}$} &
        \multicolumn{4}{c}{$\mathcal D_{FIP}^{upper}$} &
        \multicolumn{4}{c}{$\mathcal D_{GID}^{upper}$} \\
        \cmidrule(lr){2-5} \cmidrule(lr){6-9} \cmidrule(lr){10-13}
        & \textbf{Ang. (°)} & \textbf{Pos. ($cm$)} & \textbf{Mesh ($cm$)} & \textbf{Jitter ($m/s^3$) } 
        & \textbf{Ang. (°)} & \textbf{Pos. ($cm$)} & \textbf{Mesh ($cm$)} & \textbf{Jitter ($m/s^3$) } 
        & \textbf{Ang. (°)} & \textbf{Pos. ($cm$)} & \textbf{Mesh ($cm$)} & \textbf{Jitter ($m/s^3$) } \\
        \midrule

        PIP &
        77.25/\textbf{75.77} & 22.98/\textbf{21.89} & 28.89/\textbf{27.51} & 0.51/\textbf{0.88} &
        27.26/\textbf{22.78} & 12.47/\textbf{9.40} & 14.35/\textbf{11.83} & \textbf{0.28}/0.41 &
        34.35/\textbf{31.00} & 15.73/\textbf{13.41} & 17.50/\textbf{15.08} & \textbf{0.31}/0.55 \\

        PNP &
        \textbf{31.15}/33.33 & \textbf{11.84}/14.61 & \textbf{15.34}/18.98 & \textbf{0.50}/1.05 &
        \textbf{20.55}/24.02 & \textbf{7.17}/9.76 & \textbf{9.74}/14.12 & \textbf{0.28}/0.45 &
        \textbf{25.15}/25.98 & \textbf{9.44}/10.61 & \textbf{12.45}/14.43 & \textbf{0.35}/0.68 \\

        ASIP &
        \textbf{30.81}/32.22 & \textbf{11.57}/14.40 & \textbf{14.88}/18.34 & \textbf{0.22}/0.42 &
        \textbf{19.37}/25.32 & \textbf{6.46}/10.55 & \textbf{8.28}/13.84 & \textbf{0.08}/0.17 &
        \textbf{23.09}/25.83 & \textbf{7.97}/9.87  & \textbf{10.08}/12.79 & \textbf{0.10}/0.24 \\

        \bottomrule
    \end{tabular}}
    \label{tab:all_protocols_gid_smallIsBold}
\end{table*}

\begin{table*}[t]
    \centering
    \vspace{-2mm}
    \caption{\textbf{Performance comparison of garment-based MoCap methods using four key metrics.} Despite being trained on data from a single garment worn by a single user, GID achieves accuracy comparable to SOTA garment-based MoCap systems.}
    \vspace{1mm}
    \resizebox{\linewidth}{!}{
    \begin{tabular}{lcccccccccccc}
        \toprule
        \multirow{2}{*}{\textbf{Method}} &
        \multicolumn{4}{c}{$\mathcal{D}_{LIP}^{upper}$} &
        \multicolumn{4}{c}{$\mathcal{D}_{FIP}^{upper}$} &
        \multicolumn{4}{c}{$\mathcal{D}_{GID}^{upper}$} \\
        \cmidrule(lr){2-5} \cmidrule(lr){6-9} \cmidrule(lr){10-13}
        & \textbf{Ang. (°)} & \textbf{Pos. ($cm$)} & \textbf{Mesh ($cm$)} & \textbf{Jitter ($m/s^3$) } 
        & \textbf{Ang. (°)} & \textbf{Pos. ($cm$)} & \textbf{Mesh ($cm$)} & \textbf{Jitter ($m/s^3$) } 
        & \textbf{Ang. (°)} & \textbf{Pos. ($cm$)} & \textbf{Mesh ($cm$)} & \textbf{Jitter ($m/s^3$) } \\
        \midrule

        LIP~\cite{zuo2024loose} 
        & 29.84 & 11.39 & 14.33 & 0.86
        & 22.40 & 8.59 & 10.93 & 0.30
        & 23.72 & 8.65 & 10.25 & 0.51 \\

        FIP~\cite{zheng2025fip}
        & \textbf{28.24 }&\textbf{ 10.30} & \textbf{13.09} & 0.26
        & 22.77 & 8.14 & 10.57 & 0.10
        & \textbf{21.56} &\textbf{ 7.33} & \textbf{9.02} & 0.16 \\

        \textbf{GID (ASIP)}
        & 30.81 & 11.57 & 14.88 &\textbf{ 0.22}
        & \textbf{19.37} & \textbf{6.46} & \textbf{8.28} & \textbf{0.08}
        & 23.09 & 7.97 & 10.08 & \textbf{0.10} \\

        \bottomrule
    \end{tabular}
    }
    \vspace{-2mm}
    \label{tab:method_dataset_comparison}
\end{table*}






\subsection{Ablation Study}

\begin{table}[t]
    \centering
    \caption{\textbf{Ablation study of GID.} Removing FPS (Fractionized Parameter Space) trains an ASIP model directly on the single-user loose–motion paired data. Removing LSD (Location-Specific Denoiser) replaces the multi-denoiser design with a single spatiotemporal transformer. Removing ACF (Adaptive Cross-Wear Fusion) applies each denoiser independently without cross-wear fusion.}
    
    \label{tab:ablation}
    \vspace{1mm}
    \resizebox{\linewidth}{!}{
    \begin{tabular}{lccccc}
        \toprule
        & \textbf{Ang. (°)} & \textbf{Pos. ($cm$)} & \textbf{Mesh ($cm$)} & \textbf{Jitter ($m/s^3$) } & \textbf{MAE} \\
        \midrule
        w/o FPS     & 25.69 & 9.56 & 12.21 & 0.13 & --     \\
        w/o LSD     & 23.03 & 8.91 & 11.38 & 0.13 & 0.133  \\
        w/o ACF   & 24.53 & 9.80 & 12.50 & 0.13 & 0.145  \\
        \textbf{Full}        & \textbf{22.24} & \textbf{8.62} & \textbf{10.92} & \textbf{0.11} & \textbf{0.130} \\
        \bottomrule
    \end{tabular}
    }
\end{table}

As shown in the Table~\ref{tab:ablation}, the full GID model achieves the lowest global angle, position, mesh, and jitter errors (22.24°, 8.62 cm, 10.92 cm, 0.11 $m/s^3$) together with the best IMU MAE error (0.130). Collapsing the dual-stage design into a single end-to-end mapping clearly worsens all pose metrics, indicating that directly regressing from loose IMUs to pose makes optimization harder and less accurate. Using a shared denoiser instead of location-specific experts slightly degrades pose accuracy and leaves more sensor noise, showing that per-IMU experts better capture heterogeneous dynamics. The variant without the fuser yields larger angle and mesh errors and the noisiest IMU signals, highlighting the importance of cross-IMU fusion for coherent denoising. Overall, these results demonstrate that dual-stage factorization, location-specific denoising, and the fusion module are all necessary for GID to achieve the best pose and denoising performance.

\section{Live Demo}
We developed a real-time pose estimation and visualization system built on Python and Unity. After the user puts on the garment as usual, the system requires only a brief T-pose calibration (about 5 seconds) to establish initialization. Once running, the system streams IMU measurements in real time, applies calibration and normalization steps following TransPose~\cite{yang2021transpose}, and forwards the processed signals to our pose estimation model. The reconstructed motions are then rendered instantly for live visualization. In live demonstrations, the system highlights an effective balance between the \emph{comfort and natural wearability} of loose clothing and the \emph{precision} of motion capture.
It remains stable even under dynamic activities—including rapid running and jumping—showcasing strong robustness and suitability for a wide range of real-world scenarios. A demonstration video is provided in the supplementary material.
\section{Limitation and Future Work}
\paragraph{Limitation.} While denoising effectively mitigates displacement-induced perturbations, a noticeable performance gap still exists between denoised loose-wear signals and ideal tight-wear ground truth (please see supplementary materials). This indicates that current models do not yet fully capture the complex non-linear dynamics of cloth–body interactions. Moreover, our current garment designs still share a certain degree of similarity—for instance, comparable sensor placements and fabric layouts—which may limit the model’s ability to generalize across diverse clothing configurations. Integrating differentiable physics or garment simulation priors, alongside expanding the diversity of garment designs, could further reduce this gap and enable more comprehensive modeling of real-world loose sensing conditions.

\paragraph{Future Work.}
Our current experiments rely on paired loose–tight data collected from a single garment on one user. While this setup demonstrates strong data efficiency and confirms GID’s effectiveness under controlled conditions, it limits diversity in motion patterns, body shapes, and fabric types. Going forward, we aim to scale data collection and develop a denoising foundation model. This includes constructing a large, heterogeneous dataset that combines real-world recordings with physically informed synthetic data, enabling the study of scaling behavior and better cross-domain generalization. Building on this expanded corpus, we plan to train a unified model that bridges loose and tight sensing via multi-domain learning and lightweight physical priors. Ultimately, our goal is a generalizable, physically grounded foundation model for garment-based IMU denoising that fully closes the loose–tight sensing gap.
\section{Conclusion}

In this paper, we presented Garment Inertial Denoiser (GID), a lightweight and plug-and-play Transformer-based denoiser that effectively denoises IMU signals from loose-wear garments. Our key innovation is to \textit{factorize} the mapping from loose-wear IMUs to joint poses into a learnable denoising stage and a pose estimation stage. This decomposition exploits the structural similarity between loose and tight IMU signals, while location-specific denoisers capture heterogeneous motion-induced noise. Together, these design choices make GID both robust and generalizable for garment-based motion capture. To comprehensively evaluate our method and encourage future research, we further introduced GarMoCap, a comprehensive dataset comprising paired loose-pose recordings across different users, motions and garment designs. Extensive experiments demonstrate that GID, even when trained on a single garment instance, generalizes robustly across users and garment designs, significantly reducing sensor disturbances and improving downstream pose estimation accuracy. Our future research aims to train a physics-grounded desnoising foundation model in loose-wearing garment.

{
    \small
    \bibliographystyle{ieeenat_fullname}
    \bibliography{main}
}


\end{document}